# Multilingual Machine Translation with Quantum Encoder Decoder Attention-based Convolutional Variational Circuits

SUBRIT DIKSHIT[1*], RITU TIWARI[1], PRIYANK JAIN[1]

[1] Department of Computer Science and Engineering, Indian Institute of Information Technology, Pune, India

**\* Corresponding author**
**Email**: subrit@gmail.com / subritdikshit@iiitp.ac.in [1*], ritu@iiitp.ac.in[1], priyank@iiitp.ac.in[1]

**Abstract**

Cloud-based multilingual translation services like Google Translate and Microsoft Translator achieve state-of-the-art translation capabilities. These services inherently use large multilingual language models such as GRU, LSTM, BERT, GPT, T5, or similar encoder-decoder architectures with attention mechanisms as the backbone. Also, new age natural language systems, for instance ChatGPT and DeepSeek, have established huge potential in multiple tasks in natural language processing. At the same time, they also possess outstanding multilingual translation capabilities. However, these models use the classical computing realm as a backend. QEDACVC (Quantum Encoder Decoder Attention-based Convolutional Variational Circuits) is an alternate solution that explores the quantum computing realm instead of the classical computing realm to study and demonstrate multilingual machine translation. QEDACVC introduces the quantum encoder-decoder architecture that simulates and runs on quantum computing hardware via quantum convolution, quantum pooling, quantum variational circuit, and quantum attention as software alterations. QEDACVC achieves an Accuracy of 82% when trained on the OPUS dataset for English, French, German, and Hindi corpora for multilingual translations.

## 1. Introduction

In the 1940s, electronics-based computing became widespread, while the 1970s were the era of microprocessor-based classical computing. In the 2000s, artificial intelligence and deep learning-based systems became prevalent and took over the world by storm. Many modern multilingual state-of-the-art [1] networks and cloud-based translation services like Google Translate, Microsoft Translator, ChatGPT [2], DeepSeek [3] emerged and became available during this era. These Multilingual Large Language Networks are architected around Gated Recurrent Unit Networks (GRU) [4], Long Short-Term Memory (LSTM) [5], Bidirectional Encoder Representations from Transformers (BERT) [6], Generative pre-trained transformer (GPT) [7], Text-to-Text Transfer Transformer (T5) [8] and similar attention-based transformers [9] networks with finer and improved amendments to architectures. While most academicians, researchers, and organisations focused on these classical computing realm aspects and less emphasis was put on multilingual machine translation in the quantum computing realm.

Some practitioners and scholars who emphasised quantum computing for machine translation and their associated works are discussed in the Related Works section later. However, these researches under-utilized simulation and execution on quantum computing hardware along with under-exploiting the novel perceptions of quantum convolution [10], quantum pooling [11], quantum variational circuit [12] and quantum attention [13] as quantum-based software amendments that are studied, demonstrated and stunned as shortcomings in QEDACVC system.

### 1.1 Highlights

**Purpose:** The persistence drive of this research was not only to discover, examine, investigate, and compare classical multilingual state-of-the-art models and cloud-based services. Moreover, the study aims to demonstrate a path to current and future multilingual translation capability that will use the quantum computing realm (hardware, software, and model architecture) as the backbone.

**Contributions:** QEDACVC presents a quantum realm alternate solution for multilingual translations that embeds innovative operational practices, while contributions are discussed below:

- **Multilingual machine translation efficiency of 82%** – QEDACVC realises 82% of accuracy in multilingual machine translation for English, German, French, and Hindi, which can effortlessly expand to additional languages.
- **Quantum Architecture Enhancements via Quantum Convolutional Encoder Decoder augmented with Quantum Attention and Variational Circuit** – QEDACVC demonstrates successful use of the quantum encoder that consists of quantum circuit-based convolutional, pooling, and attention layers, while text data streams act as input to the quantum encoder. Quantum decoder block is made with quantum circuit-based convolutional, pooling, attention, and fully connected variational layer for interpreting the translated output data streams.
- **Multilingual representation and evaluation on OPUS corpus** – QEDACVC demonstrates training the network with OPUS [14] corpus data subsets for English, German, French, and Hindi languages.

**Challenges:** Implementation challenges for QEDACVC are discussed below:

- **Inspirational challenge** – Modern multilingual large language systems are dominated by the classical computing realm. Academicians and researchers have focused less on studying and demonstrating quantum computing implementations to experiment with multilingual translations. QEDACVC studies this gap and implements this alternate solution.
- **Limited resource availability** – QEDACVC uses open-source quantum hardware, software packages, and libraries due to a lack of funding support. The simulations are achieved on a 20-core Intel 4.7 GHz CPU, 32 GB RAM, and RTX 3070 8GB GPU hardware. Quantum experiments were executed on free quantum hardware supported by IBM and Amazon Cloud.
- **Model biases** – QEDACVC exploits open-source libraries such as Pennylane [15], Pytorch [16], Tensorflow [17], and JAX [18] as the implementation building blocks. Therefore, QEDACVC will inherit any induced biases, fluctuations, and issues that are packaged with this software and could affect QEDACVC's learning capabilities.
- **Effect of dataset imperfections** – QEDACVC was trained on multilingual data subsets for English, German, French, and Hindi from the OPUS corpus that may have inherent inconsistencies. Therefore, QEDACVC implementation is more suited to academic, research, and non-commercial applications.

**Findings:** This research work investigates and demonstrates quantum realm hardware and software for multilingual machine translation. This is achieved using the quantum encoder-decoder attention-based convolutional variational circuits.

The entire article is divided into eight segments. Section II debates the related efforts, Section III discusses the approach, Section IV with investigational outcomes, Section V as ablation studies, Section VI is the conclusion & future efforts, Section VII as declarations, and citations are registered in Segment VIII.

## 2. Related Works

This section debates the associated efforts of scholars, academicians, and works that are comparable to QEDACVC and its recital.

R Narayan et al. (2014) [19] present a machine translation system that utilizes a quantum neural network (QNN) to learn semantic patterns between Hindi and English sentences. The system analyses parts of speech and sentence structures to improve translation accuracy. Evaluation metrics such as BLEU, NIST, ROUGE-L, and METEOR scores indicate that the proposed system outperforms traditional translation methods like Google Translate and Bing Translate.

BERT (2018) [20] uses Bidirectional Encoders for linguistic illustrations. BERT is trained using unlabelled text by simultaneously conditioning each layer's left and right contexts. BERT network can be enhanced by adding one additional output layer for various applications, including question-answering and linguistic predictions, without requiring changes to the architecture. mBERT (Multilingual BERT) uses zero-shot transfer learning on morphological and syntactical tests to generate sequence illustrations for 104 languages. mBERT illustrations are either linguistic-specific or linguistic-neutral components, and the linguistic-neutral module is sufficiently all-purpose to model semantics.

UniNMT (2018) [21] boosts translation by sharing knowledge for low-resource to high-resource languages, demonstrating great results with minimal data.

Myles Doyle et al. (2020) [22] demonstrate the use of quantum computing models to perform natural language processing tasks, including language translation. The authors develop a combinational classical-quantum strategy for representing and managing corpus data using quantum circuits. Experiments conducted on IBM's quantum computers show the efficacy of the method in comparing sentence structures and meanings.

Johannes Bausch et al. (2020) [23] introduce a quantum recurrent neural network (QRNN) designed for sequence learning tasks, which are fundamental in machine translation. The QRNN unit is built using quantum neurons with parameters and amplitude amplification, enabling nonlinear activation of input polynomials. The model is evaluated on tasks like MNIST classification, demonstrating its potential in handling sequence data.

Google Neural Machine Translation System (2022) [24] utilizes a classical computing hardware and software stack to address translations. The system uses an LSTM architecture 8-8 (encoder-decoder), along with attention and residual connections for higher efficiency. The system also uses beam search to improve translation completeness while reinforcement learning boosts network BLEU scores. Google Neural Machine Translation System achieves good results on standard benchmarks and significantly reduces translation errors.

Microsoft machine translator cloud service provides an efficient mechanism for machine translation using classical computing hardware and software stack.

Akiko Eriguchi et al. (2022) [25] address the challenge of building Multilingual Neural Machine Translation systems to handle arbitrary language pairs, rather than many-to-one or bilingual setups.

mGPT [26] (2022) explores a group of autoregressive GPT-type networks with 1.3 billion hyperparameters while supporting 61 languages. The network utilizes Megatron/Deepspeed frameworks to recreate the GPT3-like network from GPT2 bases involving light-model attentions.

Mina Abbaszade et al. (2023) [27] investigate quantum methods for language translation tasks by analysing natural language on chaotic NISQ machines. The researchers employ Shannon entropy to analyse the efficiency of quantum circuits with parameters and utilize angles of rotation gates as communication means between quantum circuits of different languages. The approach adopts an idea similar to the LSTM encoder-decoder model layers, achieving promising results on a dataset of English-Persian sentence pairs.

Deepseek (2025) is an AI platform that can be used in a diverse set of natural language processing tasks and can handle multilingual data effectively for machine translation. It performs machine transitions using classical computing hardware and software stack. It uses tokenization, embeddings, multi-model architecture for training and fine-tuning.

## 3. Methodology

A formal algorithmic definition of the overall approach and the important philosophies utilized in the QEDACVC network are stated below (Table 1).

**Table 1**
Algorithmic definition of the QEDACVC network implementation.

**Algorithm 1: Overall Implementation**

**Step 0.** *Set up Environment and Libraries.*

Install the following requirement packages:

python>=3.9.12

gensim>=4.3.3, nltk>=3.9.1

jax>=0.4.30

keras>= 2.15.0, tensorflow>=2.15.0

matplotlib>=3.9.4

scikit-learn>=1.6.0

pandas>=2.2.3

PennyLane>=0.38.0, torch>=2.5.1

numpy>= 1.26.4

**Step 1.** *Load the dataset.*

Download, store, and extract the OPUS dataset.

**Step 2.** *Prepare the dataset.*

Select only English, French, Hindi, and German data subsets.

Generate and randomize a data loader for training set (size=10000), test (size=3000), and validation (size=1000) sets.

**Step 3.** *Tokenize.*

Generate a tokenizer for the inputs.

Set Batch Mode as True, Padding as maximum length, and Truncation as True.

**Step 4.** *Implement QEDACVC.*

Prepare Embeddings.

Implement sub-routines for Encoder layers with circuits of Quantum Convolutional, Quantum Pooling, and Quantum Attention.

Implement sub-routines for Decoder layers with circuits of Quantum Convolutional, Quantum Pooling, Quantum Attention, and Variational Circuit Hybrid Network.

Execute Quantum Encoder Decoder routines and build QEDACVC model layers.

**Step 5.** *Perform Training.*

Set hyperparameters - Dropout rate 0.02, epochs at 100, learning rate at 1e-5, batch size at 8, and sequence length at 64.

Perform training on QEDACVC with OPUS dataset subsets of English, French, Hindi, and German data.

**Step 6.** *Evaluate and infer from QEDACVC.*

Select and save the best instance of the QEDACVC model based on evaluation metrics – Accuracy and BLEU metric.

Generate inferences from QEDACVC.

**QEDACVC Architecture:** QEDACVC is modelled on an encoder-decoder architecture, which consists of stacked quantum circuit layers as shown in Fig. 1(a). The quantum encoder encodes the input stream of text data, while the quantum decoder decodes the output. Quantum circuit-based convolutional and quantum pooling layers work similarly to their traditional classical computing counterparts, which reduce the dimensions of the input stream with feature extractions. Quantum circuit-based attention layer allows the network to grab the most relevant predictions, while the quantum variational circuit fully connects as the last layer of the decoder. The dropout layers reduce overfitting of the QEDACVC system, and overall, Encoder / Decoder blocks are shown in Fig. 1(b). For a given input sequence $I$, $E_q$ probabilities of $q^{th}$ token element over input sequence $I$, and $D_q$ probabilities of $q^{th}$ token element over input sequence $I$, the cross-entropy loss $L_{QCVCAEDHN}$ is given by:

$$L_{QCVCAEDHN} = -\sum_q \log D_q * E_q \tag{1}$$

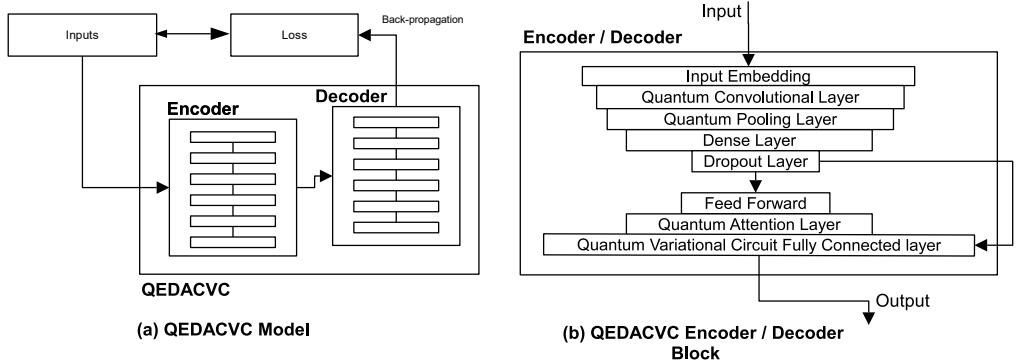

**Fig. 1.** QEDACVC Architecture.

**Quantum Convolutional and Pooling:** The Quantum Convolutional circuit-based layer allows data samples to be extracted and processed via the Convolutional circuit (Fig. 2). The circuit is initialized for 8 qubits. The two-qubit unitary consists of triple parameterized single-qubit U3 pairs, single parameterized triple Ising interaction for each pair, along with two additional U3 gates whose weights are updated during each training epoch.

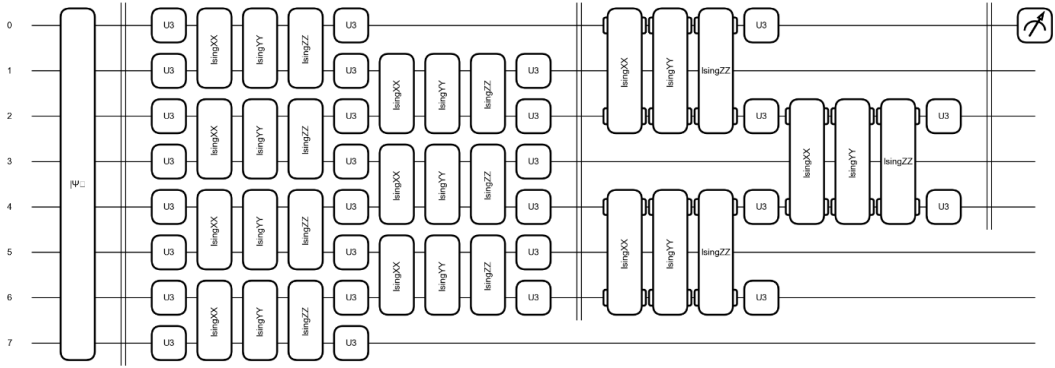

**Fig. 2.** Quantum Convolutional Circuit.

The output of the Quantum Convolutional layer is fed to a Quantum pooling circuit-based layer that performs dimensionality reduction to lower feature vectors (Fig. 3). Pooling dimensionality reduction by 50% is achieved via measurement to half on unmeasured wires of Uni-qubit conditional unitary U3 gates.

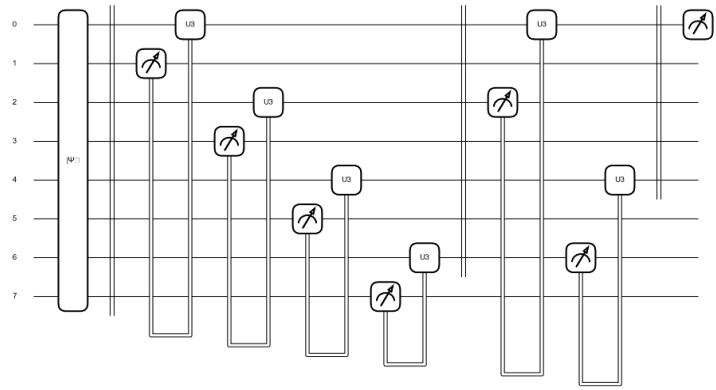

**Fig. 3.** Quantum Pooling Circuit.

The output of the Quantum pooling circuit-based layer is fed to the Quantum dense circuit-based layer as shown in (Fig. 4) with an arbitrary unitary.

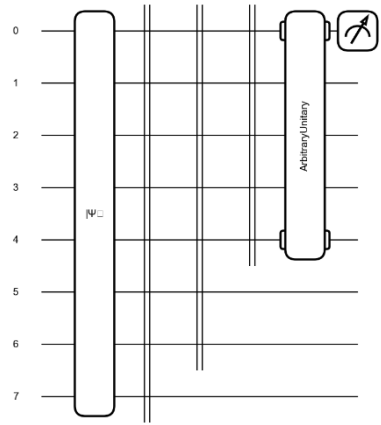

**Fig. 4.** Quantum Dense Circuit.

This complete process is depicted in (Fig. 5). The text stream of data is embedded via quantum encoding, next four four-layer quantum convolutional, and quantum pooling is applied. This latent output is condensed with a quantum dense circuit-based layer, and the final output is measured.

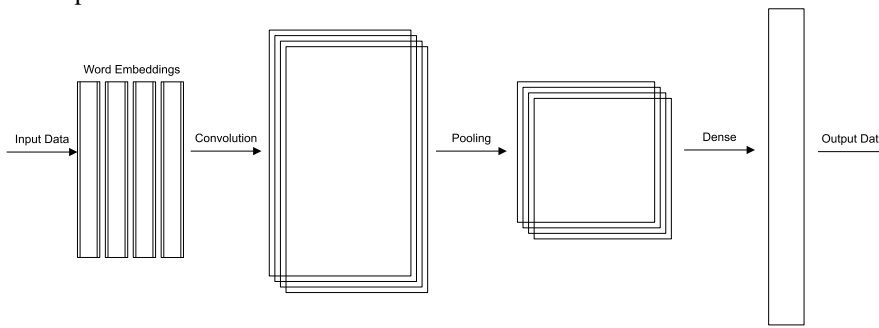

**Fig. 5.** Quantum Convolutional and Pooling.

**Quantum Attention:** QEDACVC implements a quantum circuit-based attention layer as shown in the (Fig. 6).

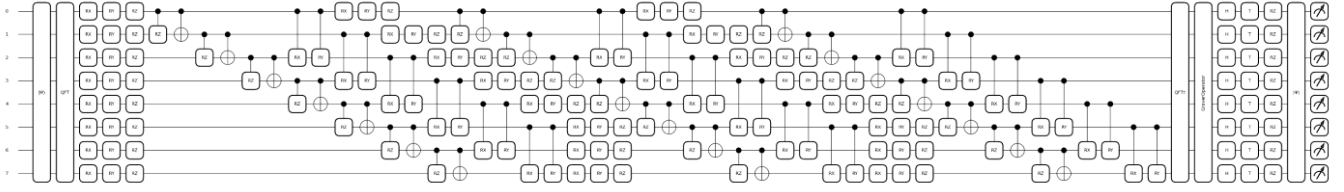

**Fig. 6.** Quantum Attention Circuit.

Quantum Attention replaces the classical attention layer (Fig. 7), where the attention score ($S_A$) for the softmax activation function, M, *A* is the ask, *B* is the key, and *C* is the value, is given by:

$$S_A(A, B, C) = M\left(\frac{AB^T}{\sqrt{d_B}}\right) C \qquad (2)$$

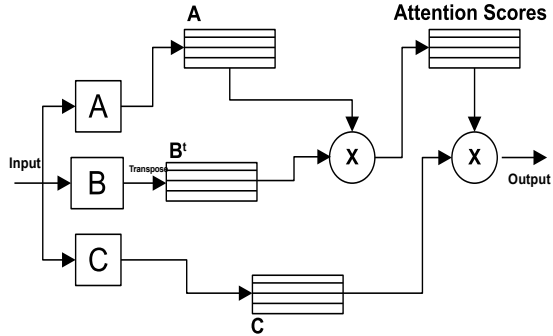

**Fig. 7.** Illustration of the Attention mechanism.

**Quantum Variational Circuit:** The Quantum Variational Circuit layer acts as fully connected layer and expands network optimization to best suit inferences (Fig. 8). The circuit is initialized for 8 qubits consisting of single-qubit Hadamard that are parametrized and rotated on Y-Axis and then shifted by CNOTs and then measured via expectation of position on Z operator. The result is a classical equivalent output of vectors, which can be further processed by the network.

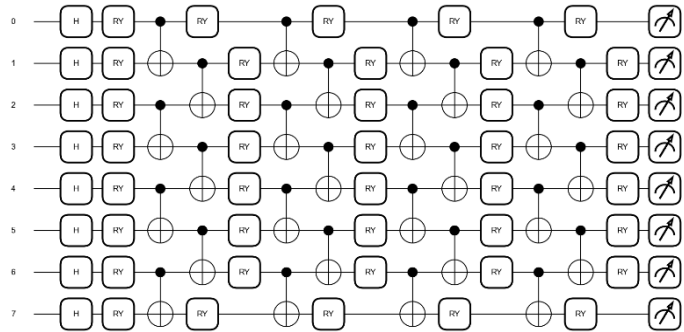

**Fig. 8.** Quantum Variational Circuit.

## 4. Experiment & Results

In this section, the experimental setting and conclusion are discussed.

**Dataset:** QEDACVC is trained and fine-tuned on the English, French, German, and Hindi language subsets from the OPUS corpus. OPUS dataset entails 1212 corpora, 58.851 trillion sentence pairs, and 747 human-spoken.

**Evaluation Strategy:** QEDACVC is evaluated on metric Accuracy [28] and BLEU Score [29].

1. **Accuracy** – *Accuracy* refers to the proportion of truthful estimates when compared with the entire examinations and expressed as:

$$Accuracy = \frac{(\alpha + \beta)}{(\alpha + \beta + \Upsilon + \delta)} \qquad (3)$$

- True positive (α): Number of right responses to predicted tokens
- True negative (β): Number of incorrect responses to predicted tokens
- False positive (ϒ): Number of predicted tokens but not in the right responses
- False negative (δ): Number of right responses, but not in predicted tokens

2. **BLEU Score** – *BLEU*, referred to as Bilingual assessment, estimates machine-translated texts and reports a stronger connection to judgments made by individuals. The BLEU score varies from 0-1, where predictions near 0 values resemblance dissimilarities and values close to 1 indicate similarities.

**Training:** Training QEDACVC is trained for a zero-shot multilingual setup. Dropout rate is set to 0.02, epochs at 50, learning rate at 2e-7, batch size 8, and maximum sequence length at 64. Table 2 and Fig. 9 display QEDACVC training and performance. Results depict that QEDACVC for English training takes around 32 steps to stabilize, whereas QEDACVC for German, French, and Hindi stabilizes around 32, 43, and 35 epochs, correspondingly.

**Table 2**
QEDACVC outcomes for training and efficiency.

| Epochs | Training Loss (By Config) [E/G/F/H] | Validation Loss (By Config) [E/G/F/H] | Performance Accuracy: BLEU Score (By Config) [E/G/F/H] |
|---|---|---|---|
| 5 | 2.114/3.372/3.197/2.787 | 2.308/3.861/2.778/2.302 | 0.328:0.535/0.312:0.479/0.306:0.398/0.301:0.491 |
| 10 | 1.098/2.022/3.060/1.447 | 0.215/2.518/2.666/0.215 | 0.397:0.542/0.366:0.486/0.312:0.425/0.388:0.498 |
| 15 | 1.006/1.500/2.911/1.327 | 0.167/1.868/2.564/0.167 | 0.459:0.587/0.391:0.500/0.320:0.477/0.400:0.539 |
| 20 | 0.760/1.294/1.511/1.002 | 0.161/1.610/0.239/0.161 | 0.533:0.601/0.435:0.522/0.388:0.489/0.423:0.552 |
| 25 | 0.465/0.907/1.385/0.613 | 0.156/1.129/0.186/0.156 | 0.623:0.628/0.452:0.573/0.448:0.510/0.413:0.576 |
| 30 | 0.187/0.623/1.012/0.247 | 0.146/0.776/0.179/0.145 | 0.699:0.643/0.512:0.600/0.519:0.523/0.487:0.591 |
| 35 | 0.113/0.563/0.581/0.149 | 0.125/0.701/0.173/0.125 | 0.702:0.734/0.588:0.632/0.607:0.597/0.491:0.674 |
| 40 | 0.040/0.458/0.216/0.053 | 0.200/0.570/0.162/0.199 | 0.734:0.767/0.648:0.688/0.651:0.643/0.523:0.705 |
| 45 | 0.031/0.314/0.136/0.040 | 0.157/0.390/0.139/0.157 | 0.762:0.782/0.700:0.700/0.685:0.734/0.550:0.719 |
| 50 | 0.015/0.024/0.017/0.019 | 0.157/0.030/0.175/0.157 | 0.818:0.892/0.719:0.801/0.728:0.791/0.584:0.819 |

**Table 2 Legends:**
English is **E**, German is **G**, French is **F**, Hindi is **H**, and the separator between language configurations is /, and : It is the separator between Accuracy and BLEU scores.

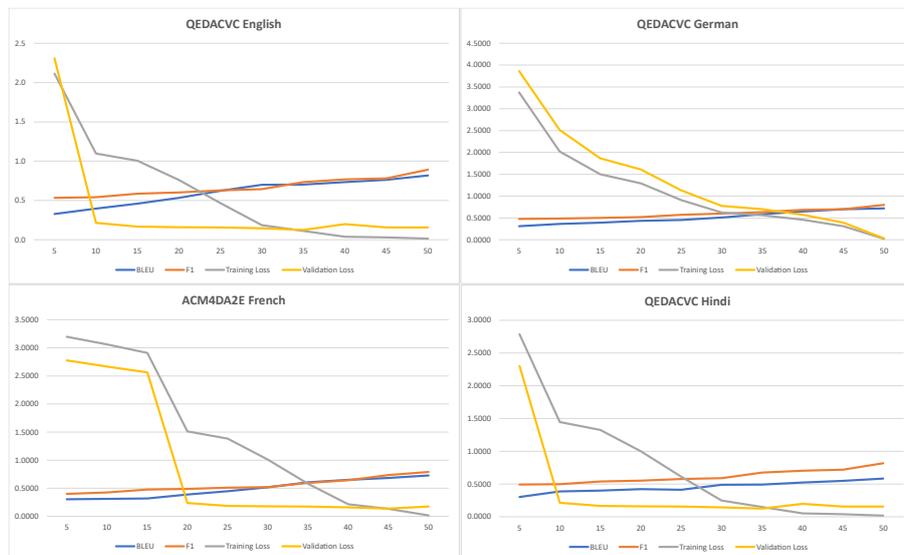

**Fig. 9.** QEDACVC configuration-based training and performance.

**Results:** The sample inferences with QEDACVC along with outputs from Google Translate, Microsoft Translator, ChatGPT, DeepSeek, GRU, LSTM, BERT, GPT, and T5 for each multilingual translation are shown in Fig. 10(a), Fig. 10(b), and Fig. 10(c).

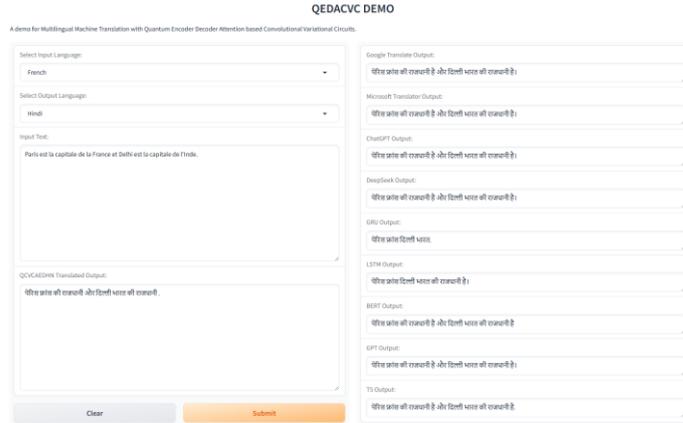

**Fig. 10 (a).** Inference with QEDACVC – French to Hindi.

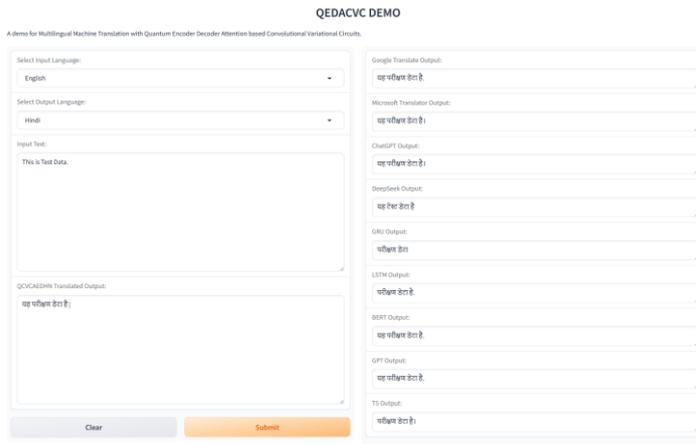

**Fig. 10 (b).** Inference with QEDACVC – English to Hindi.

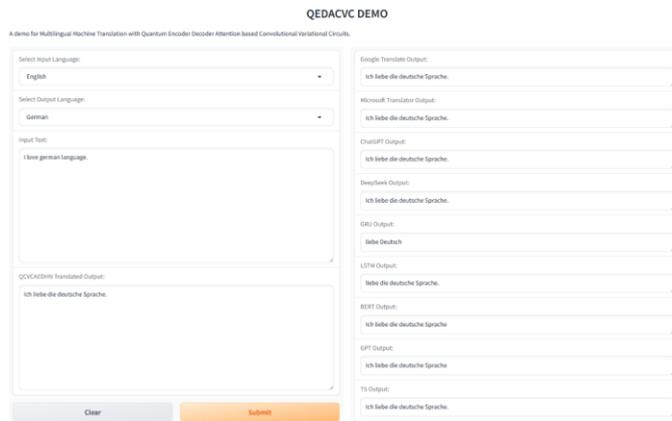

**Fig. 10 (c).** Inference with QEDACVC – English to German.

Experimental results are presented in Table 3, Table 4, and Fig. 11 for the comprehensive comparison and examination of the QEDACVC network when compared to classical multilingual large language models.

**Table 3**
Comprehensive comparison of the networks on evaluation metrics.

| | Network | English | German | French | Hindi |
|---|---|---|---|---|---|
| Accuracy / BLEU | QEDACVC | 81.8/89.2 | 71.9/80.1 | 72.8/79.1 | 58.5/81.9 |
| | GRU | 28.9/28.6 | 25.5/25.8 | 25.7/23.2 | 20.6/20.1 |
| | LSTM | 38.9/38.5 | 34.3/34.7 | 34.6/31.2 | 27.7/27.1 |
| | BERT | 82.3/81.4 | 72.5/73.3 | 73.1/65.9 | 58.6/57.2 |
| | GPT | 80.3/79.5 | 70.8/71.5 | 71.4/64.3 | 57.2/55.8 |
| | T5 | 64.4/63.7 | 56.8/57.4 | 57.2/51.6 | 45.9/44.8 |

**Table 4**
Comprehensive comparison of the network architectures and execution.

| Network Name | Total Parameters | Total Layers | Hidden Layers | Number of Heads | Training Time (Hr:Min:Sec) | Inference Speed (Sec) |
|---|---|---|---|---|---|---|
| QEDACVC | 550 | 8 | 6 | 1 | 07:05:27 | 9.26 |
| GRU | 9600 | 4 | 1 | 1 | 00:38:11 | 3.87 |
| LSTM | 50400 | 4 | 1 | 1 | 00:52:43 | 4.52 |
| BERT | 110M | 12 | 768 | 12 | 193:47:05 | 19.71 |
| GPT | 117M | 12 | 768 | 12 | 181:22:33 | 15.55 |
| T5 | 660M | 12 | 512 | 8 | 147:36:54 | 12.55 |

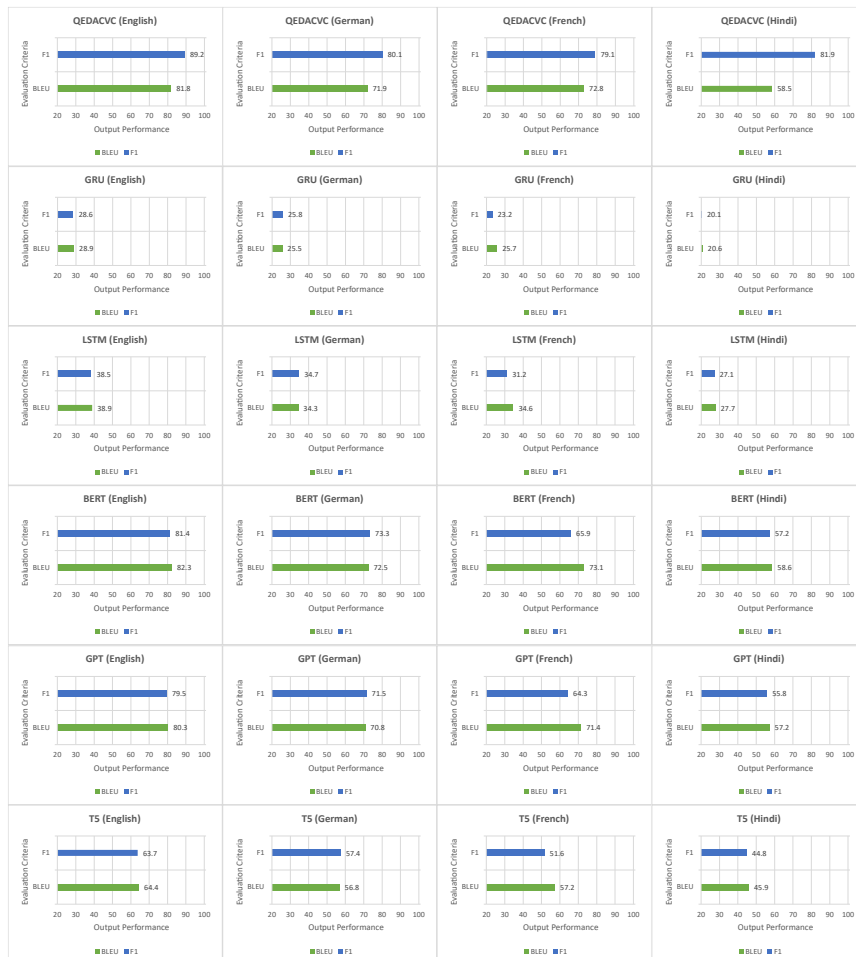

**Fig. 11.** Comprehensive comparison of QEDACVC and state-of-the-art networks.

The relative comprehensive comparison of QEDACVC with the most advanced classical natural language processing networks shows that the number of parameters plays a crucial role in accurately affecting performance. However, QEDACVC defies this limitation using the internal architectural modification techniques, which act as other major attribute that affect network performance as shown in Table 3, Table 4, and Fig. 11. QEDACVC model executes with 82% accuracy, and the network's performance is prodigious for English, followed by French, German, and then Hindi. This is due to the dataset constraints used for training the network, which can be overshadowed by adapting to other larger datasets for task-specific use cases. It is also observable from the experimental results that the QEDACVC network performs comparatively well when compared to the other state-of-the-art natural language processing systems.

## 5. Ablation Studies

This section performs experiments to observe the parameter-based effect on the performance of different QEDACVC network configurations for English, German, French, and Hindi. The different parameter-based setups are shown in Table 5, where the base model is compared with arrangements, such as the consequence of no quantum convolution layer, the consequence of adding a quantum convolution layer, the consequence of no quantum attention layer, and the consequence of adding a quantum attention layer. These results are shown in Table 6 and Fig. 12.

**Table 5**
Model Vs. Parameter-Based Performance Impact.

| Model Setup | Parameter Optimization Mode |
|---|---|
| O1 | Without Quantum Convolution Layer |
| O2 | With Quantum Convolution Layer |
| O3 | Without Quantum Attention Layer |
| O4 | With Quantum Attention Layer |
| O5 | Complete Model |

**Table 6**
Impact of parameter-based effects on the network.

| Config / Accuracy | O1 | O2 | O3 | O4 | O5 |
|---|---|---|---|---|---|
| English | 51.7 | 65.4 | 71.9 | 78.7 | 81.8 |
| German | 55.2 | 63.7 | 67.6 | 70.8 | 71.9 |
| French | 56.4 | 66.6 | 68.5 | 71.4 | 72.8 |
| Hindi | 45.5 | 48.1 | 49.8 | 56.5 | 58.5 |

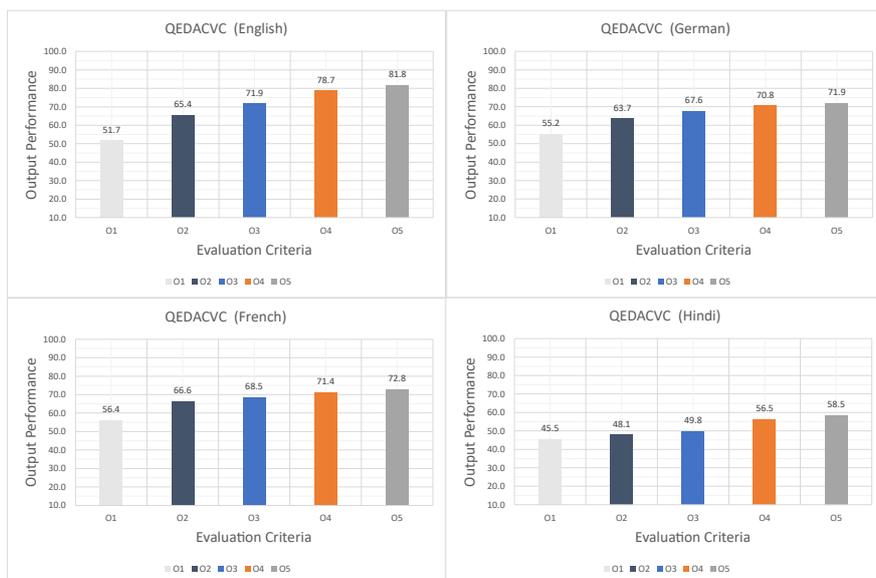

**Fig. 12.** Impact of parameter-based effects on the network.

It is evident from this demonstration that O5 achieves the best results. While at the O1 state, we have no quantum convolution layer optimization switched on, and as we introduce more and more optimizations in the network from O1 to O5, model performance increases. We also observe that model performance is best for English, followed by French, German, and then Hindi.

## 6. Conclusion & Future Work

This paper studies and discusses an effective multilingual network for the quantum realm that performs well when compared to leading state-of-the-art NLP networks running on humongous machines. The network is not 1:1 completely comparable with leading state-of-the-art NLP networks due to its size, training dataset size, and the implementation challenges that are discussed in the earlier sections. The network uses multiple forward-thinking mathematical and engineering ideas, discussed in previous segments, to accomplish this work. The reader should note that, while implementing the QEDACVC network, the capability to produce the outcomes could be slightly different on other machines. Even better results could be further achieved with modifications to the network or more efficient adaptations. The hardware utilized in this research was limited to 20 CPUs, an 8GB VRAM GPU, and 32 GB RAM, and the dataset was limited to four spoken languages from the OPUS corpus. Tweaking hyperparameter settings and quantum circuits could further optimize and achieve better inferences.

## 7. Declarations

### 7.1 Competing interests

The corresponding author declares that there is no conflict of interest on the part of all authors.

We (the authors) declare that we do not possess any competing interests to reveal, which could be financial or personal relationships with any third party that might have an impact on the article.

### 7.2 Funding Information

On behalf of all authors, the corresponding author states that this research did not receive any specific grant from funding agencies in the public, commercial, or not-for-profit sectors.

### 7.3 Author contribution

Subrit Dikshit conceptualized and designed the study, conducted the data analysis, and wrote the manuscript. Subrit Dikshit contributed to the methodology and provided critical revisions to the manuscript. Rahul Dixit, Ritu Tiwari, and Priyank Jain supervised the research, reviewed the analysis, and edited the manuscript. Subrit Dikshit assisted with data collection and contributed to the interpretation of results. All authors read and approved the final manuscript.

### 7.4 Data Availability Statement

The OPUS corpus used in this study is publicly available and can be accessed at the official repository (https://opus.nlpl.eu) of the OPUS (Open Parallel Corpus) project. All data usage complies with the respective licenses of the individual corpora within OPUS.

The dataset includes multilingual data in 747 languages. For any further inquiries or requests regarding the dataset, please refer to the OPUS official repository.

## 8. References

[1] Dikshit S, Dixit R, Shukla A., Review and analysis for state-of-the-art NLP models, International Journal of Systems, Control and Communications 15(1) (2024) 48-78, https://doi.org/10.1504/IJSCC.2024.10060461.
[2] Ray PP, ChatGPT: A comprehensive review on background, applications, key challenges, bias, ethics, limitations, and future scope, Internet of Things and Cyber-Physical Systems 3 (2023) 121-154, https://doi.org/10.1016/j.iotcps.2023.04.003.
[3] Guo D, Yang D, Zhang H, et al., DeepSeek-R1: Incentivizing Reasoning Capability in LLMs via Reinforcement Learning (2025) arXiv preprint, arXiv:2501.12948.
[4] Shen G, Tan Q, Zhang H, et al., Deep Learning with Gated Recurrent Unit Networks for Financial Sequence Predictions, Procedia Computer Science 131 (2018) 895-903, https://doi.org/10.1016/j.procs.2018.04.298.
[5] Malashin I, Tynchenko V, Gantimurov A, et al., Applications of Long Short-Term Memory (LSTM) Networks in Polymeric Sciences: A Review, Polymers (2024) 2607, https://doi.org/10.3390/polym16182607.

[6] Pires T, Schlinger E, and Garrette D. How Multilingual is Multilingual BERT? In Proceedings of the 57th Annual Meeting of the Association for Computational Linguistics, Association for Computational Linguistics (2019) 4996–5001, https://doi.org/10.18653/v1/P19-1493.

[7] Radford A, Narasimhan K, Salimans T, et al., Improving Language Understanding by Generative Pre-Training (2018) arXiv preprint, arXiv:1801.06146.

[8] Raffel C, Shazeer N, Roberts A, et al., Exploring the Limits of Transfer Learning with a Unified Text-to-Text Transformer, The Journal of Machine Learning Research 21(1) (2020) 5485–5551.

[9] Vaswani A, Shazeer N, Parmar N, et al., Attention Is All You Need, In Proceedings of the 31st International Conference on Neural Information Processing Systems (NIPS'17), New York, USA (2017) 6000–6010, arXiv:1706.03762.

[10] Henderson M, Shakya S and Pradhan S, et al., Quanvolutional Neural Networks: Powering Image Recognition with Quantum Circuits (2019) arXiv preprint, arXiv:1904.04767.

[11] Monnet M, Hanady Gebran H, Flierl AM, et al., Pooling techniques in hybrid quantum-classical convolutional neural networks (2023) arXiv preprint, arXiv:2305.05603.

[12] Stokes J, Izaac J, Killoran N, et al., Quantum Natural Gradient (2019) arXiv preprint, arXiv:1909.02108.

[13] Li G, Zhao X, Wang X, Quantum Self-Attention Neural Networks for Text Classification (2022) arXiv preprint, arXiv:2205.05625.

[14] Tiedemann J, OPUS – parallel corpora for everyone, In Proceedings of the 19th Annual Conference of the European Association for Machine Translation: Projects/Products, Riga, Latvia. Baltic Journal of Modern Computing (2016), https://aclanthology.org/2016.eamt-2.8/.

[15] Bergholm V, Izaac J, Schuld M, et al., PennyLane: Automatic differentiation of hybrid quantum-classical computations (2018) arXiv preprint, arXiv:1811.04968.

[16] Paszke A, Gross S, Massa F, et al., PyTorch: an imperative style, high-performance deep learning library, Proceedings of the 33rd International Conference on Neural Information Processing Systems, Curran Associates Inc., New York, USA, 721 (2019) 8026–8037.

[17] Abadi M, Barham P, Chen J, et al., TensorFlow: a system for large-scale machine learning, In Proceedings of the 12th USENIX conference on Operating Systems Design and Implementation (OSDI'16) USENIX Association, USA, (2016) 265–283.

[18] Bradbury J, Frostig R, Hawkins P, et al., JAX: composable transformations of Python and NumPy programs (2018) http://github.com/jax-ml/jax.

[19] Narayan R, Chakraverty S, Singh VP, Quantum neural network based machine translator for English to Hindi, Applied Soft Computing, 38 (2016) 1060-1075, https://doi.org/10.1016/j.asoc.2015.08.031.

[20] Devlin J, Chang MW, Lee K, et al., BERT: Pre-training of Deep Bidirectional Transformers for Language Understanding (2018), arXiv preprint, arXiv:1810.04805.

[21] Jiatao G, Awadalla H, Jacob D, Universal Neural Machine Translation for Extremely Low Resource Languages, Proceedings of the 2018 Conference of the North American Chapter of the Association for Computational Linguistics: Human Language Technologies, 1 (2018), https://doi.org/10.18653/v1/N18-1032.

[22] Mina A, Vahid S, Seyed M, et al., Application of Quantum Natural Language Processing for Language Translation, IEEE Access, 9 (2021) 130434-130448, https://doi.org/10.1109/ACCESS.2021.3108768.

[23] Johannes B, Recurrent Quantum Neural Networks, Advances in Neural Information Processing Systems, Curran Associates Inc., 33 (2020) 1368-1379, Recurrent Quantum Neural Networks.

[24] Wu Y, Schuster M, Chen Z, et al., Google's Neural Machine Translation System: Bridging the Gap between Human and Machine Translation (2016) arXiv preprint, arXiv:1609.08144.

[25] Eriguchi A, Xie S, Qin T, et al., Building Multilingual Machine Translation Systems That Serve Arbitrary X-Y Translations (2022) arXiv preprint, arXiv:2206.14982.

[26] Shliazhko O, Fenogenova A, Tikhonova M, et al., mGPT: Few-Shot Learners Go Multilingual (2022) arXiv preprint, arXiv:2204.07580.

[27] Abbaszade M, Zomorodi M, Salari V, et al., Toward Quantum Machine Translation of Syntactically Distinct Languages (2023) arXiv preprint, arXiv:2307.16576.

[28] Chen P, Ye J, Chen G, et al., Robustness of Accuracy Metric and its Inspirations in Learning with Noisy Labels, AAAI Conference on Artificial Intelligence (2020), https://doi.org/10.1609/aaai.v35i13.17364.

[29] Kishore P, Salim R, Todd W, et al., BLEU: a method for automatic evaluation of machine translation, Proceedings of the 40th Annual Meeting on Association for Computational Linguistics, Association for Computational Linguistics, Philadelphia, Pennsylvania, USA, (2002) 311–318, https://doi.org/10.3115/1073083.1073135.